\documentclass[letterpaper]{article} % DO NOT CHANGE THIS
\usepackage{aaai2027}  % DO NOT CHANGE THIS
\usepackage[hyphens]{url}  % DO NOT CHANGE THIS
\usepackage{graphicx} % DO NOT CHANGE THIS
\usepackage{natbib}  % DO NOT CHANGE THIS AND DO NOT ADD ANY OPTIONS TO IT
\usepackage{caption} % DO NOT CHANGE THIS AND DO NOT ADD ANY OPTIONS TO IT
\usepackage{algorithm}
\usepackage{algorithmic}

\usepackage{enumitem}
\usepackage{amsmath}
\usepackage{multirow}
\usepackage{amssymb}
\usepackage{newfloat}
\usepackage{listings}
\DeclareCaptionStyle{ruled}{labelfont=normalfont,labelsep=colon,strut=off} % DO NOT CHANGE THIS
\floatstyle{ruled}
\newfloat{listing}{tb}{lst}{}
\floatname{listing}{Listing}

\usepackage{booktabs}

\title{ControlRef: Efficient Layout-Guided Multi-Instance Generation via Anchored 4D-RoPE}
\author {
    Yunkai Yang\textsuperscript{\rm 1}, 
    Yudong Zhang\textsuperscript{\rm 2},
    Xinying Chen\textsuperscript{\rm 3},
    Haoyuan Liang\textsuperscript{\rm 2},
    Yizhuo Niu\textsuperscript{\rm 1},
    Jinshuai Cheng\textsuperscript{\rm 1},
    Kunquan Zhang\textsuperscript{\rm 1},
    Liziyue Fang\textsuperscript{\rm 1},
    Weitao Wan\textsuperscript{\rm 4},
    Runmin Dong\textsuperscript{\rm 1}\corresponding
}
\affiliations {
    \textsuperscript{\rm 1}Sun Yat-Sen University\\
    \textsuperscript{\rm 2}Tsinghua University\\
    \textsuperscript{\rm 3}Beijing Institute of Technology\\
    \textsuperscript{\rm 4}TS Martech\\
    yangyk26@mail2.sysu.edu.cn, dongrm3@mail.sysu.edu.cn
}

\nocopyright
\begin{document}

\maketitle

\begin{figure*}[h!]
  \centering
  \includegraphics[width=\textwidth]{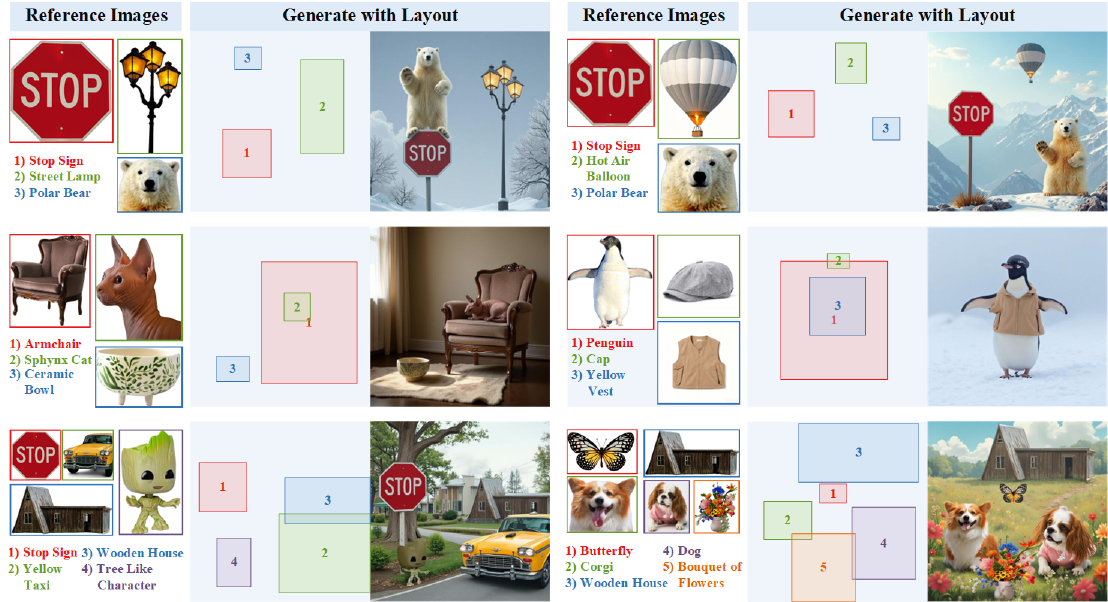}
  \caption{ControlRef resolves the spatial-frequency compromise in MM-DiTs by utilizing a novel Anchored 4D-RoPE and UILC masking to achieve highly efficient, precise layout-guided multi-instance synthesis without redundant canvas padding.}
  \label{fig:teaser}
  \vspace{-4mm}
\end{figure*}

\begin{abstract}
Layout-guided multi-instance generation is essential for controllable image synthesis in Multi-Modal Diffusion Transformers (MM-DiTs). However, integrating this capability into unified architectures remains challenging. Prior frameworks rely on redundant full-resolution canvas padding and Shifted-RoPE to manage multiple reference images. This mechanism drastically inflates computational overhead for sparse layouts and disrupts critical low-frequency RoPE features, creating a severe spatial-frequency compromise that blurs absolute spatial correspondence. To overcome these limitations, we propose ControlRef, a highly efficient and precise multi-instance synthesis framework. ControlRef utilizes a Unified Instance-Layout Control (UILC) attention mask to strictly decouple inter-instance semantic interactions and enforce precise regional binding. To further promote region-level spatial alignment, we introduce Anchored 4D-RoPE, a novel positional encoding mechanism that directly anchors tokens to their absolute geometric centers. By pre-aligning reference images to their corresponding bounding box resolutions, physically anchoring both layout and reference tokens to their absolute geometric centers, and stacking the references along the $z$-axis, Anchored 4D-RoPE natively preserves spatial priors and mitigates the spatial-frequency compromise without lossy shifting. Extensive experiments demonstrate that ControlRef achieves state-of-the-art visual fidelity and localization accuracy, while concurrently slashing inference latency by over $80\%$ in sparse layouts and reducing memory overhead by $50\%$ in dense scenarios.
\end{abstract}
\section{Introduction}
\label{sec:intro}

Diffusion Transformers (DiTs)~\cite{dit_2023} have significantly advanced high-fidelity image synthesis. Recent Multi-Modal DiTs (MM-DiTs)~\cite{mmdit_2024} process text and visual tokens within unified architectures to improve multimodal alignment. Building upon this progress, layout-guided multi-instance generation has emerged as a crucial step towards highly controllable authoring. This task requires synthesizing multiple entities that strictly conform to spatial bounding boxes and reference images.

Prior frameworks like ContextGen~\cite{contextgen_2025} achieve layout control by pasting reference images onto a full-resolution blank canvas. To process multiple references, they append additional image tokens to the sequence and prevent spatial overlap by assigning them bottom-right coordinates via Shifted 3D-RoPE~\cite{uno_2025}. However, this approach introduces notable overhead: the full-resolution canvas inflates computation for sparse layouts, while spatial shifting disrupts the low-frequency components of RoPE. This creates a spatial-frequency compromise that blurs absolute spatial correspondence. Furthermore, early architectures uniformly anchor the text modality to a static zero origin coordinate, which limits effective spatial disentanglement.

To overcome these limitations within unified architectures, we propose ControlRef, an efficient framework designed for layout-guided multi-instance synthesis. Built upon FLUX.2 [klein]~\cite{flux2_2025}, ControlRef maintains instance-level isolation by deploying a Unified Instance-Layout Control (UILC) attention mask across all transformer blocks. By restricting each reference image to its corresponding layout region, the UILC mask prevents semantic crosstalk among instances without disrupting the generative backbone's spatial understanding. Moreover, the framework accurately grounds these regions by formulating layout tokens through the concatenation of text embeddings and Fourier encoding of bounding box coordinates~\cite{gligen_2023}, enabling token-sparse layout binding. Appending reference tokens alongside these compact layout tokens reduces the overall sequence overhead, establishing a computationally efficient foundation for multi-modal conditional routing.

We introduce Anchored 4D-RoPE, a novel encoding mechanism for precise layout adherence. Instead of relying on spatial shifting or full-resolution padding, Anchored 4D-RoPE projects localized coordinates into a 4D-RoPE subspace that jointly encodes text, spatial dimensions, and multiple reference images. It aligns the resolution of each reference image to match its corresponding bounding box area. Both layout and reference tokens are anchored to their absolute geometric centers, while the multiple reference conditions are stacked within an additional subspace at discrete strides for instance separation. By using the inherent distance-dependent decay of RoPE within this geometric space, Anchored 4D-RoPE preserves absolute spatial correspondence and low-frequency features, addressing the spatial-frequency compromise of prior shifting methods. In summary, our main contributions are three-fold:

\begin{itemize}[leftmargin=*, itemsep=2pt, topsep=0pt, parsep=1pt]
\item We propose ControlRef, an efficient layout-guided multi-instance framework for DiTs, utilizing a Unified Instance-Layout Control (UILC) attention mask to enforce precise regional binding.

\item We introduce Anchored 4D-RoPE, a novel multi-axis RoPE that anchors conditional tokens to absolute geometric spaces, effectively resolving the spatial-frequency compromise of prior shifting methods.

\item ControlRef achieves SOTA visual fidelity and localization accuracy, while concurrently slashing inference latency and memory overhead by over $80\%$ and $50\%$ in sparse and dense layouts compared to prior art.
\end{itemize}

\begin{figure*}[!h]
\centering
\includegraphics[width=1.0\linewidth,height=0.6\linewidth,keepaspectratio]{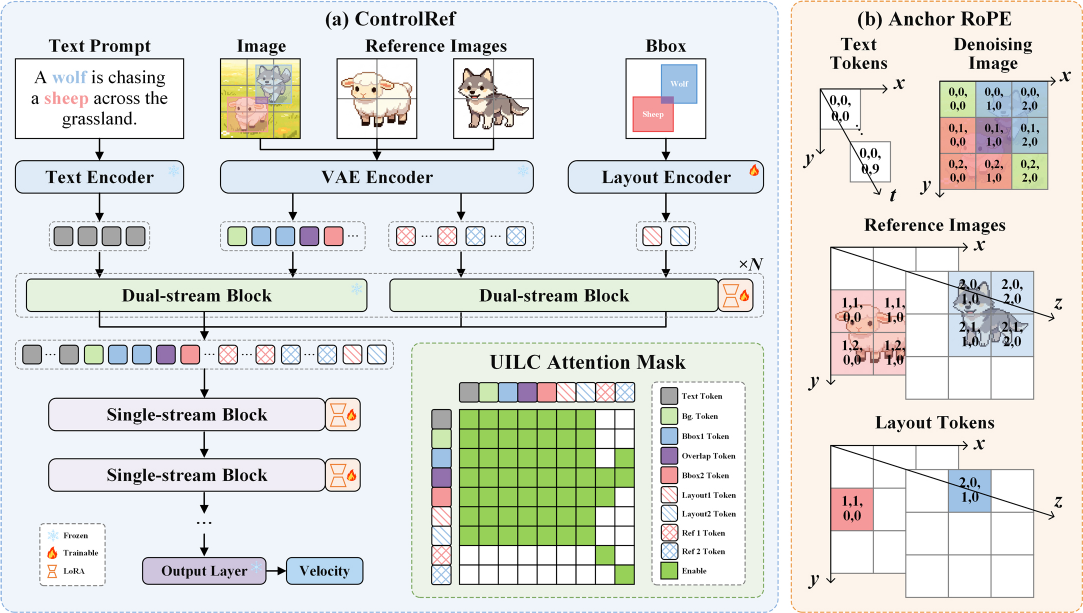}
 \caption{
 (a) Overview of ControlRef. Conditioned jointly on GLIGEN-style layout tokens and corresponding reference images, our architecture based on FLUX.2 [klein] employs the UILC attention mask to guarantee precise regional binding and instance-level isolation. (b) Anchored 4D-RoPE positional strategy. By mapping tokens to a $(z, y, x, t)$ tuple and integrating bounding box spatial offsets into the reference and layout position encodings, we establish a highly efficient spatial prior.
}
 \label{fig:2}
 \vspace{-2mm}
\end{figure*}
\section{Related Work}
\label{sec:related_work}

\subsection{Layout-to-Image Generation}

Layout-to-Image (L2I) generation~\cite{migc_2024, migcpp_2024, creatilayout_2025, layoutdiffusion_2023, layerbind_2026, groundit_2024, lawdiffusion_2023} aims to synthesize high-fidelity images that strictly conform to user-provided spatial bounding boxes. While early methods like GLIGEN~\cite{gligen_2023} enabled efficient layout-guided synthesis via Fourier feature encoding, extending this capability to modern MM-DiTs~\cite{mmdit_2024} remains challenging. Although CreatiLayout~\cite{creatilayout_2025} mitigates "modal competition" through a decoupled siamese structure, it suffers from doubled parameter overhead and poor structural scalability, hindering its extension to layout-guided multi-instance generation. Recent models, such as Z-Image~\cite{zimage_2025} and FLUX.2~\cite{flux2_2025}, increasingly rely on unified contexts and shared parameters to synchronously process text and visual tokens, maximizing multi-modal alignment natively. Moreover, prior methods typically entangle layout and text positional encodings, leading to spatial ambiguity. Maintaining the integrity of unified architectures, our approach deploys decoupled multi-axis positional encoding alongside a unified attention mask to achieve precise layout control and region binding.

\subsection{Image-to-Image Generation}

The core challenge of Image-to-Image (I2I) generation~\cite{t2i_adapter_2024, controlnet_2023, ominicontrol2_2025, unicombine_2025} lies in balancing visual fidelity to reference images with textual controllability. UNet-based frameworks~\cite{ip_adapter_2023} typically employ decoupled cross-attention to strictly separate textual and visual conditions during reference integration.

Recent MM-DiT architectures~\cite{flux_2024, flux2_2025, zimage_2025} rely on RoPE~\cite{rope_2024} to compensate for the position-agnostic nature of self-attention. While UNO~\cite{uno_2025} advocates for Shifted 3D-RoPE to integrate reference features, OmniControl~\cite{omini_2025} identifies a crucial trade-off: Shifted 3D-RoPE improves fidelity for size-mismatched signals, whereas unshifted RoPE remains superior for size-aligned inputs by maintaining precise spatial correspondence. ContextGen~\cite{contextgen_2025}, a layout-guided multi-reference generation framework, replicates a layout canvas of identical resolution to the source image and stacks it along the temporal axis. Within this temporal layer, shifted 3D-RoPE is applied to the reference image tokens to spatially decouple them from the denoising image and layout canvas. Whether utilizing Shifted 3D-RoPE for non-temporal tasks or temporal stacking for many-to-many generation (e.g., iMontage~\cite{imontage_2025}), prior shifting mechanisms merely prevent spatial overlap at the severe cost of disrupting low-frequency RoPE components. Bypassing this spatial-frequency compromise, our Anchored 4D-RoPE stacks references within a multi-axis subspace using localized positional biases, fully exploiting both high- and low-frequency features to maintain absolute spatial anchors.
\section{Methodology}
\label{sec:method}

\subsection{Preliminary}

\paragraph{Multi-axis RoPE.}

RoPE~\cite{rope_2024} encodes relative positions by partitioning $d$-dimensional components into $d/2$ pairs. By representing the $t$-th pair at position $m$ as a complex number $\mathbf{x}_m^{(t)} = x_m^{(2t)} + i x_m^{(2t+1)}$, the transformation applies a position-dependent rotation:
\begin{equation}
\tilde{\mathbf{q}}_m^{(t)} = \mathbf{q}_m^{(t)} e^{im\theta_t}, \quad \tilde{\mathbf{k}}_n^{(t)} = \mathbf{k}_n^{(t)} e^{in\theta_t},
\end{equation}
where ${\theta}_{t} = {2000}^{-2t/d}$ in practice of FLUX.2. This ensures that the attention score depends only on the relative position $m - n$: $\mathbf{Re}[\tilde{\mathbf{q}}_m^{(t)} (\tilde{\mathbf{k}}_n^{(t)})^*] = \mathbf{Re}[\mathbf{q}_m^{(t)} (\mathbf{k}_n^{(t)})^* e^{i(m-n)\theta_t}]$. The generalized Multi-axis RoPE extends this positional modeling to a $K$-dimensional coordinate tuple $\mathbf{p} = (p_1, p_2, \dots, p_K)$ by partitioning the total embedding dimension $d$ into $K$ distinct subspaces $(d_1, d_2, \dots, d_K)$, satisfying $d = \sum_{k=1}^K d_k$. By applying independent rotary transformations within each subspace, this formulation equips DiT architectures to effectively model complex multi-dimensional positional relationships while rigorously preserving relative shift-invariance.

\paragraph{FLUX.2 [klein].}

By adopting Rectified Flow~\cite{rf_2022} to linearize the transformation from noise $x_1$ to data $x_0$ via a learned velocity field $v(x_t, t)$, FLUX.2 [klein]~\cite{flux2_2025} elegantly unifies generation and editing capabilities within a single, cohesive framework. However, while recent architectures like FLUX.1~\cite{flux_2024} and Z-Image~\cite{zimage_2025} also leverage multi-axis RoPE~\cite{lumina_2024}, their spatial modeling remains constrained to 3-axis formulations. In contrast, FLUX.2 [klein] advances this paradigm by introducing a specialized 4D-RoPE variant. Furthermore, rather than employing layer-specific parameters for AdaLN~\cite{dit_2023}, FLUX.2 utilizes an efficient parameter-sharing mechanism. Specifically, within the dual-stream blocks, the text and image branches each maintain a distinct set of shared AdaLN parameters. Conversely, the single-stream blocks apply a globally unified set of AdaLN parameters across all modalities.

\subsection{The Architecture of Our ControlRef}

We propose ControlRef (Fig.~\ref{fig:2}(a)), a 4D-RoPE-empowered framework designed to further optimize computational efficiency, particularly for sparse layouts, while ContextGen~\cite{contextgen_2025} pioneers layout-guided generation by operating on full-resolution canvases. Building upon the FLUX.2 [klein] backbone and inspired by OminiControl~\cite{omini_2025}, we elegantly adapt the architecture for layout-guided multi-instance synthesis. Crucially, to ensure precise layout grounding and enhanced entity detail generation, we introduce Anchored 4D-RoPE (Fig.~\ref{fig:2}(b)), a tailored positional encoding mechanism that seamlessly projects localized instance coordinates into the unified 4D positional space $(z, y, x, t)$.

\paragraph{Conditional Tokenization.}

Following GLIGEN~\cite{gligen_2023}, we derive the layout tokens $h_l$, consisting of bounding boxes $c_{bbox} = (y_1, x_1, y_2, x_2)$ and regional text $c_{region}$, using the following formulation:

\begin{equation}
h_l = \mathbf{MLP} \left( [\phi(c_{region}), \mathbf{Fourier}(c_{bbox}) ] \right).
\end{equation}
Here, $\phi$ is the text encoder, $\mathbf{Fourier}$ denotes the Fourier spatial encoding, and $[\cdot,\cdot]$ represents concatenation along the last dimension. This concise formulation facilitates a highly efficient and seamless injection of layout information into the generative backbone. Following the default FLUX.2 [klein] configuration, reference images are VAE-encoded and appended to the sequence dimension.

\paragraph{LoRA Injection.}

To maintain minimal trainable parameters, we employ a highly selective LoRA~\cite{lora_2022}, injecting trainable matrices exclusively into the linear layers of the Attention and MLP modules. Specifically, within the dual-stream blocks, LoRA is strictly limited to the layout tokens within the text branch and the reference tokens within the denoising branch. Conversely, in the single-stream blocks, a unified, shared LoRA is utilized to process tokens across all modalities jointly.

\paragraph{Unified Instance-Layout Control Attention Mask.}

Unlike~\cite{dreamrenderer_2025}, which utilizes different attention masks across various FLUX block levels, we exploit the native multi-reference editing capabilities of FLUX.2 [klein]. This allows us to utilize a single, consistent Unified Instance-Layout Control (UILC) attention mask throughout all FLUX blocks to achieve precise regional layout binding and instance-level isolation.

Consequently, let $\mathcal{R}_{\text{img}}$ and $\mathcal{R}_{\text{txt}}$ denote the disjoint sets of indices corresponding to the denoising image tokens and text embeddings, respectively. For each instance $i$, the layout indices $\mathcal{R}_{\text{lay}_i}$ and reference image indices $\mathcal{R}_{\text{ref}_i}$ form a bijective pair, collectively representing the $i$-th entity. To regulate mutual interactions, the UILC mask is defined as:
\begin{small}
    \begin{equation}
    \mathbf{M}_{(q,k)} = 
    \begin{cases}
    \mathbf{True}, & \begin{aligned}
        &\text{if } q \in \mathcal{R}_{img} \cup \mathcal{R}_{txt} \cup \mathcal{R}_{lay} , \\[-0.5ex]
        &\phantom{\text{if }} k \in \mathcal{R}_{img} \cup \mathcal{R}_{txt} \cup \mathcal{R}_{lay},
    \end{aligned} \\[2.5ex]
    \mathbf{True}, & \begin{aligned}
        &\text{if } q \in \mathcal{R}_{\text{img}} \cap \mathbf{Mask}(\mathcal{R}_{lay_i}), \\[-0.5ex]
        &\phantom{\text{if }} k \in \mathcal{R}_{ref_i},\quad i \in [1, \dots , N],
    \end{aligned} \\[2.5ex]
    \mathbf{True}, & \text{if } q, k \in \mathcal{R}_{ref_i},\quad i \in [1, \dots , N], \\[1ex]
    \mathbf{False}, & \text{otherwise}.
    \end{cases}
    \end{equation}
\end{small}Here, $\mathcal{R}_{\text{img}} \cap \mathbf{Mask}(R_{\text{lay}_i})$ denotes the set of spatial indices within the denoising image that correspond to the $i$-th layout bounding box. The UILC mask is designed to facilitate instance-specific isolation while preserving global layout awareness. Specifically, it prevents semantic crosstalk by decoupling the interactions between disparate instances. Meanwhile, it ensures that layout information remains globally accessible, allowing the model to maintain a coherent spatial understanding.

\subsection{Anchored 4D-RoPE}
\label{sec:control_rope}

In recent FLUX-based models~\cite{uno_2025, query_kontext_2025, contextgen_2025, creatilayout_2025}, reference tokens use Shifted Multi-axis RoPE, whereas layout tokens overlap with text encodings. This spatial expansion naturally reduces attention magnitude via RoPE’s distance-dependent decay. To address this, our Anchored 4D-RoPE fully leverages the multi-axis subspace to capture spatial priors, aligning with RoPE's foundational assumption linking semantic similarity to spatial proximity. Fundamentally, the entanglement observed in previous methods stems from the original design of FLUX.1, where the native 3D-RoPE implementation remains significantly underutilized. Specifically, the text modality is uniformly anchored to a static $(0,0,0)$ coordinate irrespective of the sequence length. Consequently, the entire text representation collapses into a singular spatial point, hindering effective spatial disentanglement and leading to suboptimal layout controllability.

\begin{table*}[t]
\centering

\resizebox{\textwidth}{!}{
\begin{tabular}{l ccccc | ccccc | c}
\toprule
\multirow{2}{*}{Method} & \multicolumn{5}{c}{Fewer-Subject} & \multicolumn{5}{c}{More-Subject} & \multirow{2}{*}{AVG} \\
\cmidrule(lr){2-6} \cmidrule(lr){7-11}
& ITC & AES & IDS & IPS & AVG & ITC & AES & IDS & IPS & AVG \\
\midrule
LAMIC & 42.27 & 50.26 & 37.02 & 74.17 & 50.93 & 28.29 & 50.84 & 24.63 & 60.87 & 41.16 & 45.61 \\
XVerse & 77.65 & 53.79 & 39.47 & 71.25 & 60.54 & 43.48 & 47.68 & 15.26 & 56.12 & 40.63 & 50.29 \\
UNO & 89.86 & \textbf{58.04} & 17.53 & 75.34 & 60.19 & 77.25 & 58.90 & 7.83 & 62.94 & 51.73 & 55.58 \\
MS-Diffusion & 89.13 & 57.67 & 12.45 & 75.49 & 58.69 & 78.46 & \textbf{59.65} & 9.06 & 69.75 & 54.23 & 56.35 \\
Qwen-Image-Edit & 93.63 & \underline{57.97} & 17.71 & 73.30 & 60.65 & 86.35 & \underline{59.57} & 9.32 & 65.26 & 55.13 & 57.57 \\
OmniGen2 & \textbf{95.40} & 57.58 & 32.17 & 73.14 & 64.57 & 89.69 & 58.49 & 15.15 & 69.31 & 58.16 & 61.08 \\
Klein 9B & 90.58 & 54.12 & \underline{40.16} & 77.94 & 65.70 & 89.87 & 56.53 & 26.60 & 69.79 & 60.66 & 62.64 \\
Kontext 12B & 90.16 & 54.87 & \textbf{42.65} & 77.87 & 66.39 & \underline{90.30} & 56.08 & 27.91 & 70.93 & 61.31 & 63.33 \\
ContextGen (12B) & 92.54 & 57.50 & 35.86 & \textbf{81.23} & \underline{66.78} & 89.89 & 59.18 & \textbf{30.42} & \underline{73.35} & \textbf{63.21} & \underline{64.66} \\
\midrule
Ours (9B) & \underline{94.42} & 57.06 & 35.72 & \underline{80.74} & \textbf{66.98} & \textbf{91.14} & 58.61 & \underline{29.03} & \textbf{73.98} & \underline{63.19} & \textbf{64.72} \\
\midrule
\multicolumn{11}{c}{\textit{Closed-Source Commercial Models}} \\
\midrule
Seedream & \underline{96.92} & \textbf{59.89} & 29.43 & 78.76 & 66.25 & 95.11 & \underline{62.55} & 16.79 & 73.24 & 61.92 & 63.66 \\ 
GPT-4o & \textbf{97.63} & \underline{59.52} & 28.49 & 79.53 & 66.29 & \underline{95.37} & \textbf{62.77} & \underline{17.12} & 72.64 & \underline{61.98} & 63.71 \\
Nano Banana & 96.58 & 58.48 & \underline{34.36} & \textbf{80.87} & \textbf{67.57} & \textbf{95.48} & 60.81 & 16.67 & \textbf{74.11} & 61.77 & \underline{64.11} \\
\midrule
Ours (9B) & 94.42 & 57.06 & \textbf{35.72} & \underline{80.74} & \underline{66.98} & 91.14 & 58.61 & \textbf{29.03} & \underline{73.98} & \textbf{63.19} & \textbf{64.72} \\
\bottomrule
\end{tabular}
}
\caption{Quantitative comparison on LAMICBench++. \textbf{Bold} and \underline{underline} represent the best and second best methods.}
\label{tab:lamicpp}

\vspace{1em} 

\large 
\begin{tabular}{lccccc}
\toprule
\multirow{2}{*}{Method} & \multirow{2}{*}{Static Mem. (GB)} & \multicolumn{2}{c}{Fewer-Subject} & \multicolumn{2}{c}{More-Subject} \\
\cmidrule(lr){3-4} \cmidrule(lr){5-6}
& & Infer. Mem. (GB) & Latency (s) & Infer. Mem. (GB) & Latency (s) \\
\midrule
Kontext 12B      & 23.80 & 2.39 & 1.34 & 2.39 & 1.34 \\
Klein 9B     & 18.16 & 1.53 & 0.76 & 1.53 & 0.76 \\
ContextGen (12B)   & 27.29 & 1.50 & 1.98 & 1.66 & 2.60 \\
Ours (9B)        & 19.00 & \textbf{0.67} & \textbf{0.40} & \textbf{0.75} & \textbf{0.46} \\
\bottomrule
\end{tabular}

\caption{Efficiency comparison of different methods on Sparse and Dense layouts. Static Mem. denotes the model size, and Infer. Mem. represents the operational overhead during inference. All metrics are evaluated on a NVIDIA A100 80GB GPU.}
\label{tab:efficiency_time}

\vspace{-2mm}

\end{table*} % Comparison on LAMIC++Bench

The recently proposed FLUX.2 [klein] effectively addresses these structural deficiencies by introducing a specialized 4D-RoPE variant that jointly encodes reference, text, and denoising tokens into four distinct positional subspaces spanning a $(z, y, x, t)$ coordinate tuple. Specifically, $t$ encodes the text sequence, $(y, x)$ captures the spatial dimensions of the target image, and $z$ serves as the index for reference images, rigorously preserving relative shift-invariance. Formally, the positional indices are assigned as follows:
\begin{small}
    \begin{equation}
    \text{Pos}(t) \in 
    \begin{cases} 
    \{0\} \times \{0\} \times \{0\} \times \{0, \dots, L_{\text{txt}}-1\}, & t \in \mathcal{T}_{\text{txt}}, \\[1ex]
    \begin{aligned}
        &\{0\} \times \{0, \dots, H_{\text{img}}-1\} \\[-0.5ex]
        &\times \{0, \dots, W_{\text{img}}-1\} \times \{0\},
    \end{aligned} & t \in \mathcal{T}_{\text{img}}, \\[2.5ex]
    \begin{aligned}
        &\{t_{\text{step}} \cdot i\} \times \{0, \dots, H_{\text{ref}}^i-1\} \\[-0.5ex]
        &\times \{0, \dots, W_{\text{ref}}^i-1\} \times \{0\},
    \end{aligned} & t \in \mathcal{T}_{\text{ref}}^i.
    \end{cases}
    \end{equation}
\end{small}Here, $\mathcal{T}_{\text{txt}}$, $\mathcal{T}_{\text{img}}$, and $\mathcal{T}_{\text{ref}}^i$ denote the token sets for the text, target image, and the $i$-th reference image, respectively. The positional subspaces are formally constructed via Cartesian products of discrete coordinate sets. Specifically, $L_{\text{txt}}$ represents the text sequence length, generating the index set $\{0, \dots, L_{\text{txt}}-1\}$. Similarly, $(H_{\text{img}}, W_{\text{img}})$ and $(H_{\text{ref}}^i, W_{\text{ref}}^i)$ define the spatial dimensions of the target and reference images, yielding 2D coordinate grids such as $\{0, \dots, H_{\text{img}}-1\} \times \{0, \dots, W_{\text{img}}-1\}$. Furthermore, $t_{\text{step}}$ defines a temporal stride (defaulting to 10 in FLUX.2 [klein]) to explicitly isolate different reference instances.

Building upon this paradigm, we extend 4D-RoPE to layout tokens. We spatially anchor each layout token using its bounding box coordinate center $(\bar{y}_i, \bar{x}_i)$ discretized by patch size $P$. Combined with the instance isolation stride $t_{\text{step}}$ and the region sequence length $L_{\text{region}}^i$, the 4D positional index for a layout token $t_i \in \mathcal{T}_{\text{lay}}$ is formulated as:
\begin{small}
\begin{equation}
\begin{split}
\text{Pos}(t) \in \{ t_{\text{step}} \!\cdot\! i \} \!\times\! \{ \lfloor \frac{\bar y_i}{P} \rfloor \} \!\times\! \{ \lfloor \frac{\bar x_i}{P} \rfloor \} \\
\!\times\! \{ 0, \dots, L_{\text{region}}^i -1 \}, \quad t \in \mathcal{T}_{\text{lay}}^i.
\end{split}
\end{equation}
\end{small}Here, $\bar y_i$ and $\bar x_i$ are the bounding box centers, and $P$ denotes the patch size. The sequence $\{0, \dots, L_{\text{region}}^i-1\}$ encodes the local positional indices for the regional text, while $t_{\text{step}} \cdot i$ serves as an instance-specific offset. Leveraging RoPE’s inherent distance-dependent decay, this anchoring mechanism ensures that layout tokens naturally prioritize their corresponding local regions. By anchoring layout elements to their geometric centers $(\lfloor \bar y_i/P \rfloor, \lfloor \bar x_i/P \rfloor)$, the model simultaneously perceives their sequential text order and instance separation through the $t$ and $z$ axes, while utilizing the $(y, x)$ dimensions to enforce precise spatial binding.

To enable fine-grained detail transfer, we extend the spatial priors of Anchored 4D-RoPE to reference images. This spatially aware encoding rigorously enforces spatial isomorphism between conditions and target image. By pre-aligning the resolution of each reference image to its corresponding bounding box, we ensure an identical token count across both regions. Consequently, the 4D positional indices for the tokens of the $i$-th reference image can be formulated as:
\begin{small}
\begin{equation}
\text{Pos}(t) \in \{ t_{\text{step}} \!\cdot\! i \} \!\times\! \mathcal{H}_{\text{bbox}_i} \!\times\! \mathcal{W}_{\text{bbox}_i} \!\times\! \{0\}, \quad i \in \{1, \dots, N_{\text{lay}}\},
\end{equation}
\end{small}where $\mathcal{H}_{\text{bbox}_i}$ and $\mathcal{W}_{\text{bbox}_i}$ are the discrete sets of height and width coordinates defining the $i$-th bounding box area in the denoising image. This formulation effectively projects reference images onto a text-independent 3D spatial manifold $(z, y, x)$ to facilitate precise region-to-region correspondence. Accordingly, denoising tokens form the base of this 3D stack, with reference images layered above at $t_{\text{step}}$ intervals along the $z$-axis.

\section{Experiments}

\begin{figure*}[t]
 \centering
\includegraphics[width=1.0\linewidth,height=0.6\linewidth,keepaspectratio]{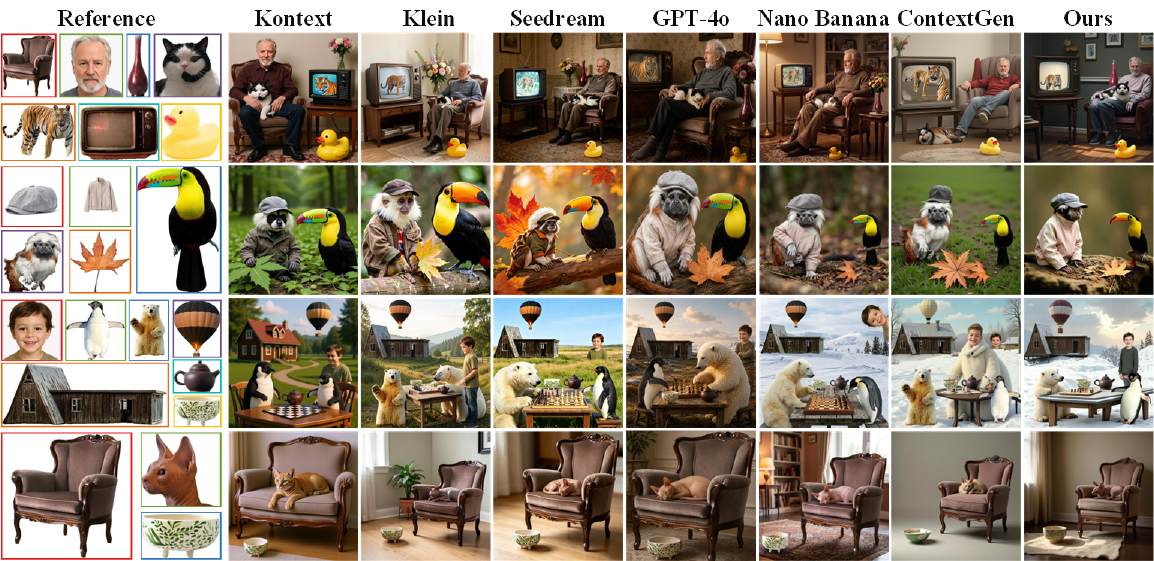}
\caption{Visual comparison on LAMICBench++.}
\label{fig:3}
\vspace{-3mm}
\end{figure*}

\subsection{Implementation Details}

We finetune FLUX.2 [klein] 9B~\cite{flux2_2025} using LoRA ($r=256$)~\cite{lora_2022} on the IMIG-100K~\cite{contextgen_2025} dataset at a $512 \times 512$ resolution. Optimization is performed via AdamW, setting the learning rate to $\eta = 1 \times 10^{-5}$ and the weight decay to $0.01$, with a total batch size of 32 on 8 NVIDIA A100 GPUs. The training pipeline comprises 8,000 SFT steps, followed by 2,000 DPO~\cite{dpo_2023} steps, strictly adhering to the preference alignment strategy established in ContextGen~\cite{contextgen_2025}.

\subsection{Datasets and Evaluation}

\paragraph{LayoutSAM-Eval.}

Our model is evaluated on a subset of 1,000 LayoutSAM-Eval~\cite{creatilayout_2025} samples containing large bounding boxes for reliable instance-level analysis. We measure instance-level attributes (spatial layout, color, texture, shape) via MLLM~\cite{minicpm_2024}, while global text-image alignment and human preference are assessed using CLIP Score and PickScore~\cite{pick_score_2023}.

\paragraph{LAMICBench++.}

For multi-subject consistency evaluation, we adopt LAMICBench++~\cite{contextgen_2025}, an aggregated benchmark containing 160 complex subject-driven generation cases. To rigorously test the scalability of our method, we evaluate across its two distinct subsets: \textit{Fewer-Subject} ($\leq 3$ images) and \textit{More-Subject} ($\geq 4$ images). Our quantitative assessment employs the four standard metrics tailored by the authors: Image-Text Consistency (ITC) via VQA~\cite{mplug_2025}, Object Preservation (IPS)~\cite{grounding_dino_2024, dinov2_2023}, Facial Identity Retention (IDS)~\cite{ids_2021}, and Aesthetic Quality (AES)~\cite{aes_2023}.

\subsection{Comparison Results}

\paragraph{Overall Quality and Instance Fidelity.} 

To evaluate generation quality and instance fidelity on LAMICBench++, we present extensive quantitative comparisons in Table~\ref{tab:lamicpp}. Our baselines encompass both open-source prior arts (LAMIC~\cite{lamic_2026}, XVerse~\cite{xverse_2026}, UNO~\cite{uno_2025}, MS-Diffusion~\cite{ms_diff_2025}, Qwen-Image-Edit~\cite{qwen_image_2025}, OmniGen2~\cite{omnigen2_2025}, Klein 9B~\cite{flux2_2025}, Kontext 12B~\cite{kontext_2025}, ContextGen~\cite{contextgen_2025}) and leading closed-source commercial models, including GPT-4o-Image, Nano Banana and Seedream 4.0~\cite{seedream_4_2025}).

Our framework demonstrates outstanding instruction adherence and instance fidelity, evidenced by achieving the highest ITC and IPS scores in the more-subject scenario. Additionally, it maintains the best overall performance (highest AVG score) in fewer-subject. Furthermore, when compared to the similarly tasked ContextGen (based on Kontext 12B), ControlRef (based on Klein 9B) demonstrates a much more pronounced improvement relative to the baseline. Specifically, ControlRef achieves substantially larger performance gains in both fewer-subject (+1.28 vs. +0.39) and more-subject (+2.53 vs. +1.90). These gains validate our framework's superior performance despite a smaller backbone. Despite falling short of commercial models in ITC and AES, ControlRef excels in IDS and IPS metrics. Ultimately, this significant advantage in instance-level control culminates in a leading overall AVG score, demonstrating superior instance fidelity.

\paragraph{Inference Efficiency and Memory Overhead.}

We evaluate our framework's efficiency on LAMICBench++. As summarized in Table \ref{tab:efficiency_time}, all metrics (measured as the average per-step latency on an NVIDIA A100 GPU in BF16 precision under identical inference steps) are compared across fewer-subject and more-subject scenarios. Focusing on inference memory and latency, ControlRef exhibits exceptional algorithmic efficiency. It not only achieves a massive $82.3\%$ reduction in latency and a $54.8\%$ decrease in inference memory compared to ContextGen (in more-subject), but also remarkably outperforms its unconditioned baseline. Specifically, ControlRef reduces the latency and memory footprint of Klein 9B by $39.5\%$ and $51.0\%$, respectively. These gains demonstrate that by replacing redundant canvas padding with our explicit spatial offset mechanism, ControlRef effectively decouples generation overhead from instance quantity, ensuring scalability in complex layouts.

\paragraph{Layout Control and Attribute Binding.}

\begin{table}[t]
\centering
\resizebox{\columnwidth}{!}{
\begin{tabular}{lcccccc}
\toprule
Method & Spatial & Shape & Color & Texture & CLIP & Pick \\
\midrule
GLIGEN             & 77.35 & 57.75 & 54.86 & 59.38 & 26.68 & 21.53 \\
LAMIC              & 77.27 & 68.74 & 69.04 & 69.96 & 23.49 & 21.91 \\
MS-Diffusion       & 85.41 & 75.21 & 73.94 & 76.08 & 26.92 & 22.22 \\
3DIS               & 88.34 & 81.30 & 80.97 & 82.52 & 26.75 & 21.89 \\
CreatiLayout       & 93.59 & 78.89 & 77.43 & 79.62 & \textbf{27.99} & 22.44 \\
MIGC               & 86.04 & 73.37 & 71.07 & 74.88 & 25.50 & 21.10 \\
EliGen             & \underline{94.05} & 87.01 & 83.84 & 87.31 & 26.89 & 22.27 \\
ContextGen         & 93.96 & \underline{88.36} & \underline{87.44} & \textbf{89.26} & 27.26 & \textbf{22.47} \\
\midrule
Ours               & \textbf{94.52} & \textbf{89.18} & \textbf{88.76} & \underline{89.05} & \underline{27.93} & \underline{22.45} \\
\bottomrule
\end{tabular}
}
\caption{LayoutSAM-Eval Quantitative Results.}
\label{tab:layoutsam_eval}
\vspace{-3mm}
\end{table}

We evaluate our layout control against leading methods on LayoutSAM-Eval, including GLIGEN~\cite{gligen_2023}, LAMIC~\cite{lamic_2026}, MS-Diffusion~\cite{ms_diff_2025}, 3DIS~\cite{3dis_2024}, CreatiLayout~\cite{creatilayout_2025}, MIGC~\cite{migc_2024}, EliGen~\cite{eligen_2025} and ContextGen~\cite{contextgen_2025}. Table~\ref{tab:layoutsam_eval} presents the quantitative results on LayoutSAM-Eval benchmark. ControlRef achieves the highest spatial alignment score among the evaluated methods, indicating the effectiveness of the Anchored 4D-RoPE mechanism in regional localization. In terms of instance fidelity, our approach outperforms most baselines in shape and color preservation, and performs comparably to the leading methods in texture synthesis. These results suggest that the UILC attention mask helps isolate instances and mitigate semantic leakage. Finally, the evaluations on CLIP Score and PickScore show that the framework maintains overall visual quality while adhering to the layout constraints.

\subsection{Ablation Study}

\paragraph{Effect of UILC Attention Mask Across FLUX Blocks.}

We evaluate the performance of ControlRef across different block-level configurations on LAMICBench++ to determine the optimal integration of the UILC attention mask. The evaluated configurations include a baseline without the mask, injection only in dual-stream blocks, injection only in single-stream blocks, and injection across all blocks. Table~\ref{tab:uilc_mask} shows that incorporating the mask throughout all blocks yields the highest average score, driven by peak performance in image-text consistency and aesthetic quality. While the single-stream block configuration approaches the performance of the full integration, it exhibits a slight drop in overall fidelity. In contrast, applying the mask exclusively to dual-stream blocks underperforms the baseline across most metrics. This indicates that an incomplete application fails to ensure sufficient instance-level isolation and permits attribute leakage. The strategy of integrating the mask across all blocks ensures that spatial and attribute constraints are enforced consistently throughout the entire architecture, maximizing global aesthetic alignment and instance-level precision.

\begin{table}[htbp]
\centering
\begin{tabular}{lccccc}
\toprule
Method & ITC & AES & IDS & IPS & AVG \\
\midrule
w/o Mask            & 88.35 & 54.70 & \textbf{31.93} & 73.96 & 62.24 \\
Dual Blocks         & 88.54 & 53.91 & 28.54 & 72.89 & 60.97 \\
Single Blocks       & \underline{91.23} & \underline{56.99} & 31.28 & \textbf{76.61} & \underline{64.03} \\
All Blocks          & \textbf{92.73} & \textbf{57.74} & \underline{31.86} & \underline{76.53} & \textbf{64.72} \\
\bottomrule
\end{tabular}
\caption{Ablation Study on UILC Injection Strategy.}
\label{tab:uilc_mask}
\vspace{-1mm}
\end{table}

\begin{table}[h]
\centering
\resizebox{\columnwidth}{!}{
\begin{tabular}{lcccccc}
\toprule
Method & Spatial & Shape & Color & Texture & Pick \\
\midrule
w/o Ref Offset              & 92.95 & 86.18 & 86.22 & 88.45 & 21.97 \\
w/o Layout Offset           & \underline{94.01} & 87.40 & 86.09 & 88.68 & 22.03 \\
w/o Layout Tokens           & 93.83 & \underline{88.24} & \underline{87.48} & \textbf{89.54} & \textbf{22.51} \\
ControlRef                  & \textbf{94.52} & \textbf{89.18} & \textbf{88.76} & \underline{89.05} & \underline{22.45} \\
\bottomrule
\end{tabular}
}
\caption{Ablation Study on Anchored 4D-RoPE Positional Index.}
\label{tab:anchor_rope}
\vspace{-3mm}
\end{table}

\paragraph{Analysis of Anchor-RoPE Positional Index.}

We conduct ablation studies on the LayoutSAM-Eval benchmark to evaluate the layout binding capability of the Anchored 4D-RoPE mechanism, with results presented in Table~\ref{tab:anchor_rope}. Removing reference or layout offsets results in degraded spatial precision. This indicates that injecting absolute bounding box offsets into both reference images and layout tokens is necessary for accurate spatial grounding. Otherwise, the model maps visual features with positional ambiguity. The variant omitting layout tokens, which relies solely on reference offsets, exhibits a drop in spatial alignment. This highlights the role of layout tokens in explicitly modeling inter-instance relationships within complex configurations. The full ControlRef framework achieves the most balanced performance across all evaluated metrics. By explicitly encoding geometry, Anchored 4D-RoPE limits the interference between layout constraints and instance-level features, effectively decoupling spatial layouts from visual attributes.
\section{Conclusion}

In this paper, we introduce ControlRef, a novel framework designed to address the challenges of layout-guided multi-instance generation. By decoupling spatial configurations from visual attribute binding, our approach effectively mitigates attribute leakage and instance fidelity degradation common in existing models. Specifically, the Anchored 4D-RoPE mechanism uses absolute coordinate offsets for precise spatial grounding, while the UILC attention mask enforces instance-level isolation during synthesis. Experiments on the LayoutSAM-Eval and LAMICBench++ benchmarks demonstrate that ControlRef achieves leading performance in spatial controllability and instance fidelity. Furthermore, the framework substantially improves computational efficiency, requiring lower inference latency and a smaller memory footprint than prior spatial padding methods. Future work includes extending these layout-binding mechanisms to video generation tasks and exploring fine-grained attribute routing within unified transformer architectures.

% \appendix
% xxx

% \section*{Acknowledgments}
% xxx

\bibliography{aaai2027}


\clearpage
\setcounter{page}{1}

\appendix

\renewcommand{\thesection}{\Alph{section}}
\setcounter{section}{0}
\setcounter{figure}{0}
\setcounter{table}{0}

\twocolumn[
    \begin{center}
        \LARGE\bfseries Appendix
        % \LARGE\bfseries Supplementary Material
    \end{center}
    \vspace{2mm}
]

\section{More Implementation Details}

\paragraph{More Details on UILC attention mask.}

Compared to its predecessor, FLUX.1.Kontext, FLUX.2 [klein] introduces support for multi-reference image conditioning. Regarding AdaLN modulation, the reference images are assigned a fixed timestep of 0 (denoting the noise-free data distribution) and share the identical AdaLN linear projection layer with the denoising image. Owing to the time-invariance of the reference timestep, FLUX.2 [klein] inherently accommodates native KV Cache optimization through an asymmetric attention routing mechanism. Specifically, the reference image tokens exclusively undergo self-attention, while their KV states are concatenated with those of the textual and denoising branches, enabling the model to effectively capture multi-reference contexts for guided image editing.

\begin{figure}[!h]
\centering
\includegraphics[width=0.9\linewidth,height=0.7\linewidth,keepaspectratio]{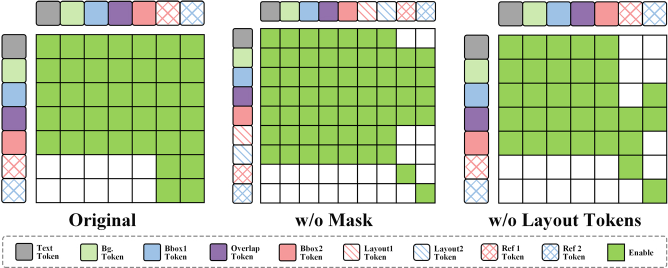}
 \caption{The attention mask settings in ablation experiments.}
 \label{fig:appx_1}
\end{figure}

To preserve the native KV Cache optimization, we assign all reference images a fixed timestep of $t=0$, while maintaining the architectural consistency illustrated in Fig.~\ref{fig:appx_1} across all ablation studies involving the UILC attention mask. Concretely, the reference tokens exclusively undergo intra-branch self-attention, whereas the textual and denoising queries are allowed to attend to the cached KV states of the reference context. Furthermore, regarding the isolation masks among multiple reference images, we adopt the standard practice of prior multi-reference generation DiT-based frameworks~\cite{contextgen_2025, easycontrol_2025, dreamrenderer_2025}. This masking strategy serves a dual purpose: it strictly precludes information leakage between distinct instances and significantly reduces the computational overhead of the attention mechanism.

\paragraph{Generative Backbone Selection.}
Our proposed ControlRef framework features a model-agnostic architecture, theoretically allowing any model compatible with the anchored 4D-RoPE design to serve as its backbone. Currently, within the open-source community, 4D-RoPE is uniquely adopted by the FLUX.2 series. Accordingly, we evaluate our approach on both FLUX.2 [klein] 4B and FLUX.2 [klein] 9B to demonstrate its efficacy and scalability. As detailed in Table \ref{tab:backbone_selection}, we compare our method with both the foundational FLUX.2 backbones and prior leading architectures, namely the FLUX.1-based Kontext 12B and its advanced variant ContextGen. Quantitative results indicate that ControlRef yields a more substantial performance lift over its baselines. Specifically, Ours (4B) and Ours (9B) achieve absolute improvements of +2.20 and +2.08 in the AVG metric over Klein 4B and 9B, respectively, significantly outpacing the marginal +1.33 gain that ContextGen achieves over its Kontext 12B baseline. Crucially, ControlRef establishes a new state-of-the-art (SOTA) with exceptional parameter efficiency. Ours (9B) attains the highest overall performance while requiring considerably fewer parameters than the 12B-based ContextGen. Furthermore, even our lightweight 4B variant remarkably surpasses the Kontext 12B baseline, underscoring the architectural superiority of our design.

\begin{table}[t]
\centering
\begin{tabular}{lccccc}
\toprule
Method & ITC & AES & IDS & IPS & AVG \\
\midrule
Klein 4B       & 89.84 & 54.28 & 31.91 & 72.15 & 61.85 \\
Klein 9B       & 90.21 & 55.16 & 32.33 & 72.86 & 62.64 \\
Kontext 12B    & 90.30 & 55.40 & 34.12 & 73.53 & 63.33 \\
ContextGen     & 91.38 & \textbf{58.24} & \textbf{32.72} & \underline{76.32} & \underline{64.66} \\
\midrule
Ours (4B)       & \underline{91.70} & 57.07 & \underline{32.09} & 75.35 & 64.05 \\
Ours (9B)       & \textbf{92.73} & \underline{57.74} & 31.86 & \textbf{76.53} & \textbf{64.72} \\
\midrule
\end{tabular}
\caption{Ablation Study on different generative backbone.}
\label{tab:backbone_selection}
\end{table}

\paragraph{Prompt Details for Image Generation.} The underlying FLUX.2 backbone functions as a unified architecture for both image editing and generation. Consequently, when conditioned on multiple reference images, it can accept either editing-oriented or generation-oriented prompt templates. Through preliminary evaluations on the vanilla FLUX.2 [klein] 9B within a multi-reference generation paradigm, we empirically observe that editing-based templates yield notably superior generation quality. Motivated by this finding, our proposed framework explicitly adopts editing-oriented prompt templates during the training and inference phase. Our editing-based prompt template was used: ``Generate an image described as `\{GENERATIVE PROMPT\}' by arranging the given reference images. While preserving the original layout, key object features and human identities (including facial details), adjust element poses for better composition and naturally fill in the background. Ensure the result appears natural and visually harmonious.''

\section{More Method Details}

\paragraph{DPO for Diffusion Models.} Recently, Reinforcement Learning from Human Feedback (RLHF) has emerged as a pivotal paradigm for aligning generative models with human preferences. While standard supervised training establishes foundational generation capabilities, RL-based optimization enables direct fine-tuning toward complex, non-differentiable objectives, such as perceptual aesthetics, prompt adherence, and structural fidelity, which traditional likelihood-based losses fail to explicitly capture. While the Supervised Fine-Tuning stage successfully equips the model with layout-conditioned reference placement capabilities, it inadvertently introduces a copy-and-paste hacking problem. Because reference images in the training set are directly extracted from the target ground-truth images, the model easily degenerates into this trivial hacking shortcut rather than performing true context-aware generation. To address this limitation, following ContextGen~\cite{contextgen_2025}, we leverage Diffusion-DPO~\cite{diffusion_dpo_2024} during the post-training stage to refine text-visual alignment and align with human preferences. This transition enables the model to evolve from mastering simple collage-like compositions to synthesizing complex, harmoniously integrated multi-instance scenes.

Formally, RLHF aims to optimize a conditional distribution $p_\theta(x_0|c)$ over conditioning contexts $c \sim \mathcal{D}_c$. The objective is to maximize the expected reward $r(c, x_0)$ while penalizing the Kullback-Leibler (KL) divergence relative to a pre-trained reference distribution $p_{\text{ref}}(x_0|c)$:
\begin{equation}
\begin{aligned}
&\max_{\theta} \mathbb{E}_{c \sim \mathcal{D}_c, x_0 \sim p_\theta(x_0|c)} [r(c, x_0)] \\
&\quad - \beta D_{\text{KL}} \left( p_\theta(x_0|c) \parallel p_{\text{ref}}(x_0|c) \right),
\end{aligned}
\label{eq:rlhf_objective}
\end{equation}
where $\beta$ is a hyperparameter governing the strength of KL regularization to prevent policy drift.

Directly solving Eq. (\ref{eq:rlhf_objective}) via reinforcement learning, most notably PPO~\cite{ppo_2017}, suffers from severe instability and heavy computational overhead due to the requirement of fitting an explicit reward model. DPO~\cite{dpo_2023} circumvent these challenges by reparameterizing the reward function. Mathematically, the constrained optimization problem in Eq. (\ref{eq:rlhf_objective}) yields an analytical closed-form solution for the optimal policy $p^*_\theta$:
\begin{equation}
p^*_\theta(x_0|c) = \frac{1}{Z(c)} p_{\text{ref}}(x_0|c) \exp \left( \frac{1}{\beta} r(c, x_0) \right),
\label{eq:optimal_policy}
\end{equation}
where $Z(c) = \int p_{\text{ref}}(x_0\vert{}c) \exp \left( \frac{1}{\beta} r(c, x_0) \right) dx_0$ is the partition function. Rearranging Eq. (\ref{eq:optimal_policy}) allows us to express the ground-truth reward function $r(c, x_0)$ strictly in terms of the optimal policy and reference distribution:
\begin{equation}
r(c, x_0) = \beta \log \frac{p^*_\theta(x_0|c)}{p_{\text{ref}}(x_0|c)} + \beta \log Z(c).
\label{eq:reward_reparam}
\end{equation}

For preference optimization, we assume access to a pairwise preference dataset $\mathcal{D} = \{(c, x_0^w, x_0^l)\}$, where $x_0^w$ and $x_0^l$ denote the preferred (winner) and dispreferred (loser) outputs given context $c$, human preferences are commonly modeled via the Bradley-Terry model: $P(x_0^w \succ x_0^l \vert{} c) = \sigma \left( r(c, x_0^w) - r(c, x_0^l) \right)$, where $\sigma(\cdot)$ is the sigmoid function. By substituting the reparameterized reward from Eq. (\ref{eq:reward_reparam}) into the Bradley-Terry model, the partition function $Z(c)$ conveniently cancels out. Consequently, the parameterized policy $p_\theta$ can be directly optimized using the negative log-likelihood loss:
\begin{equation}
\begin{gathered}
\mathcal{L}_{\text{DPO}}(\theta) = -\mathbb{E}_{(c, x_0^w, x_0^l) \sim \mathcal{D}} \bigg[ \log \sigma \Big( \beta \log \frac{p_\theta(x_0^w|c)}{p_{\text{ref}}(x_0^w|c)} \\
- \beta \log \frac{p_\theta(x_0^l|c)}{p_{\text{ref}}(x_0^l|c)} \Big) \bigg].
\end{gathered}
\label{eq:dpo_loss}
\end{equation}

While the objective in Eq. (\ref{eq:dpo_loss}) operates on explicit log-likelihoods in autoregressive models, evaluating $p_\theta(x_0\vert{}c)$ in diffusion models is computationally intractable due to the high-dimensional marginalization over latent trajectories. Following Diffusion-DPO, we reparameterize the log-likelihood ratio using the Evidence Lower Bound, which simplifies the preference optimization into a weighted difference of noise prediction errors across diffusion timesteps $t \sim \mathcal{U}(0, 1)$. Specifically, given a condition $c$, comprising prompt and layout constraints, noise $\epsilon \sim \mathcal{N}(0, \mathbf{I})$, and noisy latents $x_t = (1 - t) * x_0 + t * x_1$, the Diffusion-DPO training objective is formulated as:
\begin{equation}
\begin{aligned}
\mathcal{L}_{\text{Diff-DPO}}(\theta) = & -\mathbb{E}_{\mathcal{D}, t, \epsilon} \bigg[ \log \sigma \Big( - \frac{\beta}{2} \big( \|\epsilon^w - \epsilon_\theta(x_t^w, t, c)\|_2^2 \\
& - \|\epsilon^w - \epsilon_{\text{ref}}(x_t^w, t, c)\|_2^2 \\
& - (\|\epsilon^l - \epsilon_\theta(x_t^l, t, c)\|_2^2 \\
& - \|\epsilon^l - \epsilon_{\text{ref}}(x_t^l, t, c)\|_2^2) \big) \Big) \bigg],
\end{aligned}
\label{eq:diffusion_dpo}
\end{equation}
where $\beta$ is a hyperparameter controlling the trade-off between preference alignment and regularization against the reference model. Following the empirical validation in ContextGen~\cite{contextgen_2025}, we set $\beta = 1000$ as the optimal trade-off value to ensure stable preference optimization. The preference dataset conforming to Diffusion-DPO is constructed by designating the ground-truth image as the winner sample $x_0^w$. Correspondingly, the loser sample $x_0^l$ is constructed by naively pasting the multi-reference image crops onto a blank canvas according to their specified layout bounding boxes. This construction explicitly penalizes rigid layout stitching and encourages the model to generate seamless visual transitions.

\section{More Quantitative Experiments}

\begin{figure*}[!h]
\centering
\includegraphics[width=0.9\linewidth,height=0.7\linewidth,keepaspectratio]{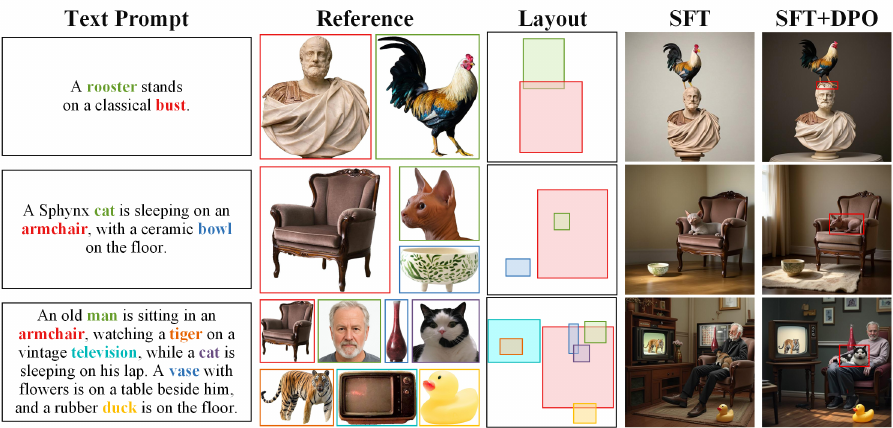}
 \caption{Visual comparison of DPO Strategy.}
 \label{fig:appx_2}
\end{figure*}

\paragraph{Improvement of Reinforcement Learning.} To evaluate the effectiveness of DPO post-training strategy, we present an ablation study in Table~\ref{tab:dpo_ablation}. For a fair comparison, all baseline and proposed variants are trained under identical LoRA rank configurations and $\beta$. As observed, integrating Diffusion-DPO brings consistent and remarkable improvements in both aesthetic quality and image-text consistency. Specifically, applying DPO to our framework boosts the AES score by +2.11 points and enhances the ITC score by +1.40 points. A similar upward trend is observed in ContextGen, where DPO elevates AES and ITC by +2.91 and +4.19, respectively. These empirical results demonstrate that while SFT effectively establishes structural layout generation capabilities, preference optimization via DPO significantly aligns the model with fine-grained visual aesthetics and semantic prompt adherence, ultimately driving our final model to achieve the highest overall average score.

Figure~\ref{fig:appx_2} presents qualitative comparisons that highlight the perceptual enhancements brought by preference alignment. As illustrated, models post-trained with DPO exhibit substantially smoother transitions and seamless blending at object boundaries and intersections, effectively removing boundary artifacts observed in the SFT baseline. Furthermore, in challenging multi-reference scenarios, DPO demonstrates a superior capability in preserving entity consistency for small-scale objects, faithfully maintaining their fine-grained visual features and structural identity across generations. These visual refinements confirm that DPO successfully steers the generation trajectory toward human perceptual preferences, yielding more visually pleasing and contextually harmonized outputs.

\subsection{Additional Results on COCO-MIG}

Due to space limitations in the main manuscript, we provide extended quantitative experiments on COCO-MIG~\cite{migc_2024} to further validate the superiority of our framework in spatial layout binding and attribute grounding.

\paragraph{Benchmark and Metrics.}

Derived from the established COCO~\cite{coco_common_2014} distribution, COCO-MIG serves as a challenging layout-to-image benchmark comprising 800 rigorously color-annotated instances, specifically curated to penalize spatial ambiguity and cross-object interference. To thoroughly dissect model performance across distinct generation facets, spatial and attribute accuracy are quantitatively measured via global and instance-level Success Rates (SR and I-SR, relying on precise mIoU computations and strict color correctness), while semantic alignment is captured by both global and local CLIP scores (G-C and L-C).

\begin{table}[t]
\centering
\resizebox{\columnwidth}{!}{
\begin{tabular}{lccccc}
\toprule
Method & ITC & AES & IDS & IPS & AVG \\
\midrule
ContextGen (SFT)     & 86.84  & 54.19 & \underline{32.37} & \underline{76.78} & 62.55 \\
ContextGen (SFT+DPO) & 91.03  & \underline{57.10} & 26.83 & 75.71 & 62.67 \\
Ours (SFT)           & \underline{91.33}  & 55.63 & \textbf{32.79} & \textbf{76.95} & \underline{64.17} \\
Ours (SFT+DPO)       & \textbf{92.73} & \textbf{57.74} & 31.86 & 76.53 & \textbf{64.72} \\
\midrule
\end{tabular}
}
\caption{Ablation Study on DPO Strategy.}
\label{tab:dpo_ablation}
\end{table}

\begin{table}[t]
\centering
\resizebox{\columnwidth}{!}{
\begin{tabular}{lccccc}
\toprule
Method & SR & I-SR & mIoU & G-C & L-C \\
\midrule
GLIGEN             & 4.25  & 29.56 & 27.44 & 25.21 & 20.90 \\
LAMIC              & 1.25  & 13.56 & 21.17 & 21.82 & 18.71 \\
MS-Diffusion       & 4.50  & 28.22 & 34.69 & 25.50 & 20.77 \\
3DIS               & 18.88 & 55.44 & 49.35 & 23.72 & 20.40 \\
CreatiLayout       & 19.12 & 54.69 & 48.96 & \textbf{26.22} & 20.70 \\
MIGC               & 27.75 & 66.44 & 56.96 & \underline{26.21} & 21.47 \\
EliGen             & 26.00 & 64.12 & 59.23 & 24.92 & 20.58 \\
ContextGen         & \textbf{33.12} & \underline{69.72} & \underline{65.12} & 25.86 & \underline{21.87} \\
\midrule
Ours               & \underline{32.76} & \textbf{69.91} & \textbf{65.23} & 26.12 & \textbf{21.95} \\
\bottomrule
\end{tabular}
}
\caption{COCO-MIG Quantitative Results.}
\label{tab:coco_mig}
\vspace{-2mm}
\end{table}

\paragraph{Quantitative Comparisons.}

We evaluate our layout binding against leading methods on COCO-MIG, including GLIGEN~\cite{gligen_2023}, LAMIC~\cite{lamic_2026}, MS-Diffusion~\cite{ms_diff_2025}, 3DIS~\cite{3dis_2024}, CreatiLayout~\cite{creatilayout_2025}, MIGC~\cite{migc_2024}, EliGen~\cite{eligen_2025}, and ContextGen~\cite{contextgen_2025}.

As presented in Table~\ref{tab:coco_mig}, our framework establishes new sota performance across key spatial control and attribute binding metrics. Specifically, our method achieves top-ranked performance on instance-level success rate, spatial overlap precision, and localized text-visual consistency, while maintaining a highly competitive second-place standing on global success rate. Compared to the strongest existing baseline ContextGen, our approach yields consistent gains in fine-grained instance grounding and layout accuracy, alongside superior local prompt alignment. Furthermore, when evaluated against earlier region-aware diffusion baselines, our model exhibits decisive performance leads across all structural indicators. These quantitative results confirm that our framework successfully mitigates attribute leakage and boundary blurriness, ensuring highly faithful spatial layout binding in multi-instance synthesis scenarios.

\end{document}